\documentclass[10pt,twocolumn,letterpaper]{article}

\usepackage{iccv}
\usepackage{times}
\usepackage{epsfig}
\usepackage{graphicx}
\usepackage{amsmath}
\usepackage{amssymb}
\usepackage[titletoc, toc, page]{appendix}
\graphicspath{ {./images/} }

\usepackage[breaklinks=true,bookmarks=false]{hyperref}

\iccvfinalcopy 

\ificcvfinal\fi

\begin{document}

\title{Open Problems in Computer Vision for Wilderness SAR and The Search for Patricia Wu-Murad}

\author{Thomas Manzini, Dr. Robin Murphy\\
Texas A\&M University\\
400 Bizzell St, College Station, TX 77843\\
{\tt\small tmanzini@tamu.edu, \tt\small robin.r.murphy@tamu.edu}}

\maketitle
\ificcvfinal\thispagestyle{empty}\fi

\begin{abstract}

    This paper details the challenges in applying two computer vision systems, an EfficientDET supervised learning model and the unsupervised RX spectral classifier, to 98.9 GB of drone imagery from the Wu-Murad wilderness search and rescue (WSAR) effort in Japan and identifies 3 directions for future research.
    There have been at least 19 proposed approaches and 3 datasets aimed at locating missing persons in drone imagery, but only 3 approaches (2 unsupervised and 1 of an unknown structure) are referenced in the literature as having been used in an actual WSAR operation.
    Of these proposed approaches, the EfficientDET architecture and the unsupervised spectral RX classifier were selected as the most appropriate for this setting. 
    The EfficientDET model was applied to the HERIDAL dataset and despite achieving performance that is statistically equivalent to the state-of-the-art, the model fails to translate to the real world in terms of false positives (e.g., identifying tree limbs and rocks as people), and false negatives (e.g., failing to identify members of the search team). 
    The poor results in practice for algorithms that showed good results on datasets suggest 3 areas of future research: more realistic datasets for wilderness SAR, computer vision models that are capable of seamlessly handling the variety of imagery that can be collected during actual WSAR operations, and better alignment on performance measures.

\end{abstract}

\section{Introduction}

Patricia Wu-Murad is a 60-year-old female hiker who went missing along the Komado Komo trail in southern Nara, Japan on April 10, 2023. 
The search for her included hundreds of wilderness search and rescue (WSAR) personnel, several of which operated small unmanned aerial systems (sUAS) that collected 98.9 GB of imagery.
\begin{figure}[h!]
\centering
\includegraphics[width=70mm,scale=0.5]{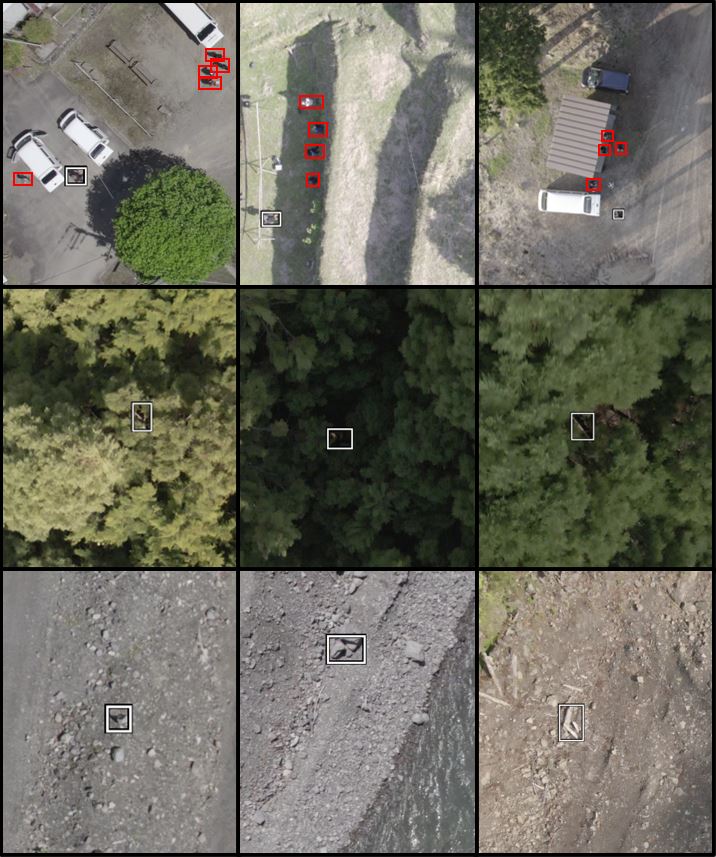}
\caption{Predictive failures from \textit{Tiled\_EfficientDET$_{174}$}. Row 1: Failure to find members of the search team (false negatives in red). Row 2: Misclassification of tree limbs as people. Row 3: Misclassification of rocks as people. Errors corresponding to rows 2 and 3 were particularly common in the data collected in the field.}
\label{fig:model_fails}
\end{figure}
This volume of imagery can be difficult to inspect manually, so these operations were augmented by using CV techniques to identify specific regions of interest that may be used to help focus search efforts.
This paper details the efforts and challenges in developing and deploying CV systems in an actual WSAR operation and highlights 3 future directions for research. At the time of writing Patricia Wu-Murad remains missing\footnote{Imagery, code, models, and instructions for replication can be found at \url{https://github.com/CRASAR/WiSAR/}}.

\section{Wilderness SAR Operations}
WSAR is a class of field operations that involves searching for and rescuing a missing person in austere wilderness settings. 
This is done by sending teams of people to the areas where the missing person is suspected to be located where they will perform search activities such as ground search patterns (e.g., grid-search), following scent trails using dogs, or flying drones to collect aerial imagery to uncover information that will lead to the missing person's location.

\subsection{sUAS for Wilderness SAR}
\label{suas_wsar}
sUAS are small, portable aerial robots with onboard sensors and cameras that can be controlled by a ground crew to search an area for the missing person or signs of their presence \cite{mayer2019drones}.
In WSAR settings, these sUAS have become valuable as they enable ground teams to collect high resolution imagery over a large area.
These sUAS have been widely used in WSAR \cite{mcrae2019using, grogan2018use} and have resulted in the rescue of hundreds of missing persons \cite{djidrm}.

sUAS operations for WSAR can take several forms.
For example, hasty searches are flown at several hundred feet above the ground in a boustrophedon or lawnmower pattern collecting nadir imagery at fixed intervals.
In some situations, flight plans can be designed that attempt to maximize the probability of detection of a person based on lost person behavior models \cite{williams2020collaborative}.
Other approaches consider flight paths for fixed wing vehicles that allow the collection of multiple views of objects without the ability to hover \cite{bashyam2019uavs}.

Even though the operators of the sUAS are typically able to inspect the aerial view while in flight, they are usually capturing high definition imagery that can be further analyzed, a process colloquially referred to as ``squinting."
It is common for search teams to spend hours observing the imagery in an effort to ensure that no clues or regions of interest were missed.

\subsection{Computer Vision for Wilderness SAR}
\label{sec:cv_sar}
In supporting WSAR operations, the efforts of the CV community have been primarily focused on automating the squinting process described in Section \ref{suas_wsar}.

On the unsupervised side, several color-based approaches have been proposed \cite{Morse2012ColorAD, marshall2015color, niedzielski2017nested}, and one histogram-based approach was discussed in \cite{agcayazi2016resquad}.

On the supervised learning side, three datasets dominate the literature: the SARD dataset \cite{sambolek2021automatic}, the HERIDAL dataset \cite{maruvsic2018region}, and most recently, the WiSARD dataset, which explores both electro-optical and infrared modalities \cite{broyles2022wisard}.
These three datasets focus on person detection in wilderness settings and are annotated using bounding boxes.

At least 19 models have been trained on these datasets, and all of them lean heavily on the existing literature in object detection. 
\begin{figure}[h!]
\centering
\includegraphics[width=80mm,scale=0.5]{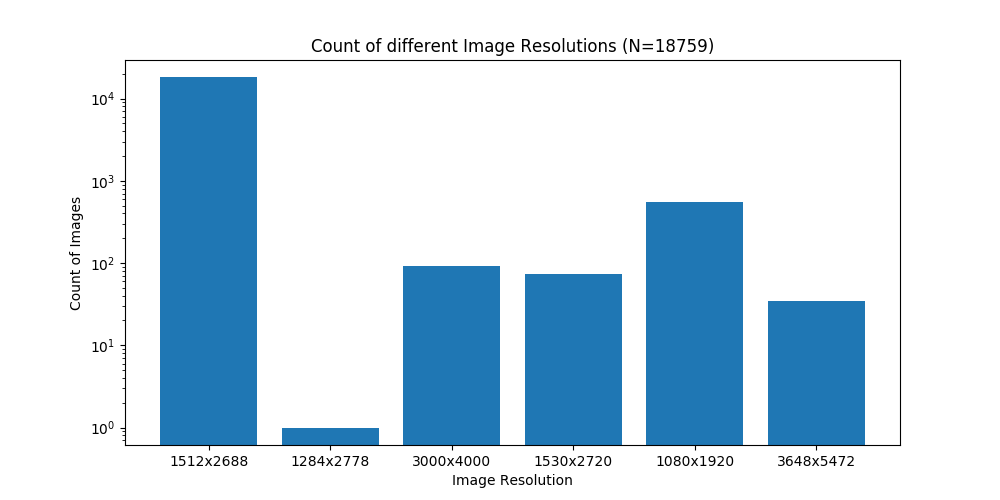}
\caption{The count of each distinct image resolution in the sampled frames collected during the search for Patricia Wu-Murad.}
\label{fig:resolutions}
\end{figure}
The YOLO architecture \cite{redmon2016you} is by far the most prolific in this space as multiple variants have been trained and evaluated on the SARD dataset \cite{sambolek2021automatic, betti2023yolo, caputo2021human, bachir2022investigating, han2022ehdnet}, the HERIDAL dataset \cite{caputo2021human, popa2022real}, and the WiSARD dataset \cite{broyles2022wisard}.
The Single Shot Detection architecture \cite{liu2016ssd} and the FasterRCNN \cite{ren2015faster} architectures have been explored in \cite{kundid2020multimodel} and \cite{maruvsic2018region, sambolek2020person}, respectively. 
The RetinaNet architecture \cite{lin2017focal} was also extensively experimented with in \cite{pyrro2021rethinking, pyrro2022air}.
The EfficientDET architecture was explored in \cite{dousai2022detecting}; however, they did so on an unpublished subset of cropped HERIDAL samples, making it difficult to compare its effectiveness.

Outside of approaches driven by these datasets, other authors explore supervised CV systems to find missing persons in avalanches \cite{bejiga2017convolutional} and search for missing persons in maritime settings\cite{sahana2022person, lygouras2019unsupervised}, along side other applications \cite{bernal2023hierarchically}. 

Based on a review of the literature, at least two unsupervised methods have been used in a real world search and rescue operation: The RX spectral classifier described in \cite{Morse2012ColorAD, marshall2015color} was used in the Texas Memorial Day Floods in 2015 \cite{proft2015spectral}, and the commercial product Loc8 \cite{loc8_software, blackburn2021detectability}, a system which relies on color based methods, has also seen operational use \cite{loc8_use}. 
The commercial product SARUAV \cite{saruav_software} has also been used operationally and claims to use a deep neural network, though it does not provide details on training data or model architecture \cite{niedzielski2021first}.

\section{Drone Operations \& Data Details}
\label{sec:ops_data}
The search for Patricia Wu-Murad, generated 98.9 GB of imagery over approximately 3 weeks. 
This data consisted of 128 high resolution photos and 243 high resolution videos at different resolutions as shown in Figure \ref{fig:resolutions}.

Teams primarily flew two types of missions: high altitude nadir imagery and low altitude oblique imagery.
High altitude nadir imagery is the standard in WSAR operations; however, as tree cover in this region of Japan is dense, drone teams began collecting low altitude oblique imagery below the treeline.
Thus high altitude nadir imagery was reserved for open areas such as river beds, farmland, and trails.

\begin{figure}[h!]
\centering
\includegraphics[width=75mm,scale=0.5]{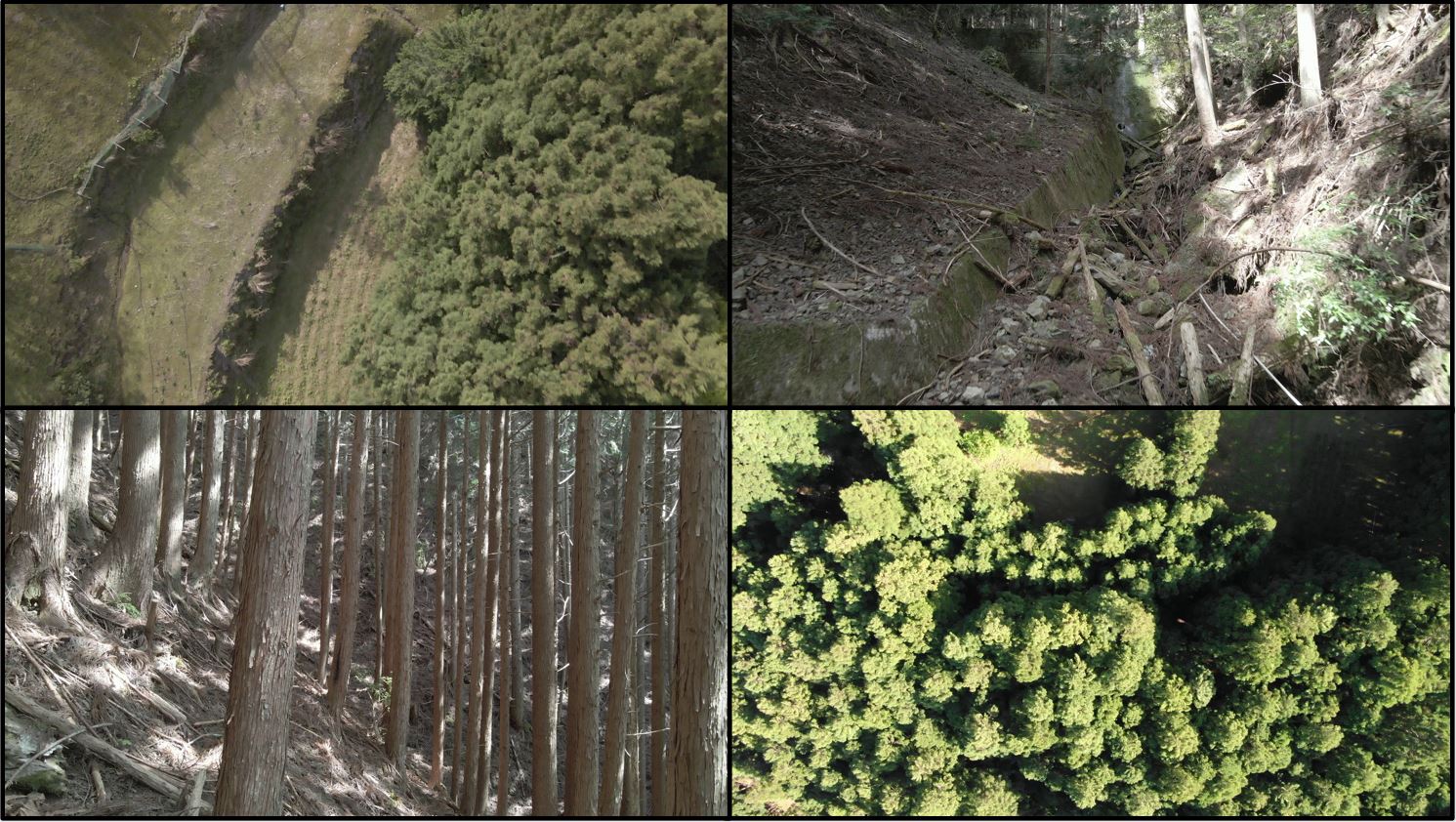}
\caption{Sample images selected from the 18,759 frames analyzed during the search for Patricia Wu-Murad.}
\label{fig:sample_wu}
\end{figure}

In order to process this data in a reasonable amount of time, all videos were sampled at a rate of 2 frames per second, and all images were processed as is. 
This generated a set of 18,759 images to be analyzed. 
A sample of these photos can be seen in Figure \ref{fig:sample_wu}.

It is worth noting that because these flights were conducted over a mountainous region and at varying altitudes and resolutions there is substantial scale variance in potential regions of interest and pixels per inch.

\section{Computer Vision Techniques Employed}

As discussed in Section \ref{sec:ops_data}, the imagery consisted of both oblique and nadir views. 
Efforts were directed to the high altitude, mostly nadir imagery, because the low altitude oblique imagery was collected such that any potential region of interest be immediately evident, and the high altitude imagery is the most time intensive to review.

Both a supervised and unsupervised method were employed to support the search efforts: an augmented version of the RX spectral classifier detailed in \cite{Morse2012ColorAD, proft2015spectral}, and an EfficientDet bounding box model \cite{tan2020efficientdet, effdet} trained on the HERIDAL dataset.
These two methods are discussed herein.

\subsection{Unsupervised Approach}
The RX classifier was employed because one of the ground WSAR teams had already utilized the Loc8 software on this imagery and because the RX classifier had already seen use in previous WSAR field operations.
As discussed in \cite{proft2015spectral}, this method selects anomalous pixels in an image based on the mahalanobis distance function and a given p-value resulting in almost all images having some set of anomalous pixels. 
This would mean that the output of this classifier must be inspected for every image.
However, manual inspection of the 18,759 images was not practical.
As a result, additional post processing steps were performed in order to further refine the classifier outputs. 
DBSCAN clustering was performed on the detected anomalous pixels in order to group anomalies together.
Photos that had $<$1 and $>$4 clusters and clusters that contained $<$209 pixels were discarded.

This approach resulted in 5234 candidate images that were all manually reviewed.
Following that review, none of the candidate images were elevated to the search team.\footnote{Hyperparameters and sample outputs are included in Appendix \ref{app:rx}.}

\subsection{Supervised Approach}
This section details the building and the performance of a supervised model to support the ongoing WSAR activities.

\subsubsection{Choice of Dataset}
The 3 principal datasets referenced in the literature were considered: SARD, HERIDAL, and WiSARD. 
SARD was discarded as its altitude, view angles, and target class poses were too different from the data that had been collected in the field and the inspection tasks required in this effort.
The WiSARD dataset was also not pursued as it has only recently been released and has not been evaluated except by its publishing authors.
Instead, the HERIDAL dataset was utilized as it partially aligned with the imagery inspection tasks at hand and it had been studied academically.

\subsubsection{Choice of Model Architecture}
As detailed in Section \ref{sec:cv_sar}, the most explored network architectures in this space are YOLO and RetinaNet. 
The RetinaNet-based AIR model architecture detailed in \cite{pyrro2021rethinking, pyrro2022air} was considered, but not pursued, due to concerns about replicability stemming from the sensitivity to initial network conditions referenced in \cite{pyrro2022air}.
Following this, the YOLO \cite{jocher2022ultralytics} and EfficientDET \cite{tan2021efficientnetv2} architectures were explored and training on the HERIDAL dataset began.

Given the active WSAR operations, the YOLOv5 approach was abandoned because poor validation set performance was observed compared to the EfficientDET model.

The model pursued was a 512x512 EfficientDET model and training samples were generated by selecting appropriately sized crops of from the HERIDAL dataset.\footnote{The details of this model training process are included in Appendix \ref{app:ed_training}.}
Inference was then performed by regularly spacing 512x512 tiles from high resolution imagery as described in \cite{pyrro2022air}.
All predicted bounding boxes with overlap are merged into a new bounding box surrounding the overlapping boxes.

Two versions of this model are reported here: \textit{Tiled\_EfficientDET$_{84}$} was run as a hasty effort to collect results to share with search teams, and \textit{Tiled\_EfficientDET$_{174}$} was run following further training.
The number that follows the model name denotes the number of epochs that the model was trained for.
Both models were run on the imagery collected in the field. 

Following the termination of search efforts, these models were evaluated on the HERIDAL test set and the results are reported in Table \ref{tab:perf}. 
\textit{Tiled\_EffecientDET$_{174}$} receives an average precision score of 91.5, which is statistically equivalent to the current SOTA model performance (p$\leq$0.02), 
without tuning hyperparameters on the test set as described in \cite{pyrro2022air}. 

\begin{table}[]
\begin{tabular}{lccc|}
\cline{2-4}
\multicolumn{1}{l|}{}                                & \multicolumn{1}{c|}{P}    & \multicolumn{1}{c|}{R}    & AP                    \\ \hline
\multicolumn{4}{|c|}{VOC2012 Evaluation \cite{pascal-voc-2012}}                                                                                             \\ \hline
\multicolumn{1}{|l|}{Mean Shift Segmentation \cite{turic2010two, pyrro2022air}} & \multicolumn{1}{c|}{18.7} & \multicolumn{1}{c|}{74.7} & -                     \\
\multicolumn{1}{|l|}{Saliency Guided VGG16 \cite{bovzic2019deep}}          & \multicolumn{1}{c|}{34.8} & \multicolumn{1}{c|}{88.9} & -                     \\
\multicolumn{1}{|l|}{Faster R-CNN (2019) \cite{bovzic2019deep}}            & \multicolumn{1}{c|}{58.1} & \multicolumn{1}{c|}{85}   & -                     \\
\multicolumn{1}{|l|}{Faster R-CNN (2018) \cite{maruvsic2018region}}            & \multicolumn{1}{c|}{67.3} & \multicolumn{1}{c|}{88.3} & \textbf{86.1}                  \\
\multicolumn{1}{|l|}{Two-stage Multimodel CNN \cite{kundid2020multimodel}}       & \multicolumn{1}{c|}{68.9} & \multicolumn{1}{c|}{94.7} & -                     \\
\multicolumn{1}{|l|}{SSD \cite{kundid2020multimodel}}                            & \multicolumn{1}{c|}{4.3}  & \multicolumn{1}{c|}{94.4} & -                     \\
\multicolumn{1}{|l|}{YOLOv5s \cite{caputo2021human}}                        & \multicolumn{1}{c|}{75.3} & \multicolumn{1}{c|}{69.4} & 73.1                  \\
\multicolumn{1}{|l|}{YOLOv5m \cite{caputo2021human}}                        & \multicolumn{1}{c|}{79.7} & \multicolumn{1}{c|}{81.2} & 81                    \\
\multicolumn{1}{|l|}{AIR with NMS** \cite{pyrro2022air}}                 & \multicolumn{1}{c|}{90.1} & \multicolumn{1}{c|}{86.1} & 84.6                  \\
\multicolumn{1}{|l|}{Tiled\_EffecientDET$_{84}$ (ours)}            & \multicolumn{1}{l|}{58.0}     & \multicolumn{1}{l|}{53.7}     &  \multicolumn{1}{l|}{49.3}                     \\
\multicolumn{1}{|l|}{Tiled\_EffecientDET$_{174}$ (ours)}            & \multicolumn{1}{l|}{78.6}     & \multicolumn{1}{l|}{84.0}     &  \multicolumn{1}{l|}{80.1}                     \\ \hline
\multicolumn{4}{|c|}{SAR-APD Evaluation \cite{pyrro2022air}}                                                                                             \\ \hline
\multicolumn{1}{|l|}{AIR with NMS** \cite{pyrro2022air}}                 & \multicolumn{1}{c|}{90.5} & \multicolumn{1}{c|}{87.8} & 86.5                  \\
\multicolumn{1}{|l|}{AIR with MOB** \cite{pyrro2022air}}                 & \multicolumn{1}{c|}{94.9} & \multicolumn{1}{c|}{92.9} & \textbf{91.7}                  \\
\multicolumn{1}{|l|}{Tiled\_EffecientDET$_{84}$ (ours)}            & \multicolumn{1}{l|}{65.3}     & \multicolumn{1}{l|}{61.4}     & \multicolumn{1}{l|}{61.3} \\
\multicolumn{1}{|l|}{Tiled\_EffecientDET$_{174}$ (ours)}            & \multicolumn{1}{l|}{82.8}     & \multicolumn{1}{l|}{91.4}     & \multicolumn{1}{l|}{\textbf{91.5}} \\ \hline
\end{tabular}
\newline
\caption{\label{tab:perf}Precision (P), Recall (R), and Average Precision (AP) on the HERIDAL test set. 
Entries marked in bold are either SOTA performance or are within the 95\% confidence interval.
Precision and Recall scores are not bolded because of inconsistencies in the literature regarding thresholds.
Precision and Recall scores reported for \textit{Tiled\_EffecientDET} models are reported with a 0.5 threshold.
Entries marked with ** were attained following hyperparameter tuning on the test set.}
\end{table}

\subsubsection{Inference Details \& Model Failures}

Inference of the EfficientDET models on the 18,759 images was performed. \textit{Tiled\_EffecientDET$_{84}$} generated bounding boxes in 1522 images at 50\% confidence, and \textit{Tiled\_EffecientDET$_{174}$} generated bounding boxes in 1377 images at 60\% confidence, all of which were manually reviewed.
This process resulted in 4 candidate images that were elevated to the search teams\footnote{These 4 candidate images are included in Appendix \ref{app:cand}.}.

Unfortunately, these models generate a substantial number of false positives and negatives, routinely failing to recognize search team members and frequently mislabeling bushes, rocks, and branches as people as shown in Figure \ref{fig:model_fails}.
Finally, chromatic aberrations near photo edges are occasionally labeled as people, presumably because of red and green coloration.


\section{Future Directions}
The current body of academic literature leaves much to be desired when it comes to CV applications for WSAR.
There are three directions that are needed to better refine this area of work: more realistic and diverse datasets, models which leverage the data that is available in practice, and community alignment on evaluation metrics.

\textbf{More Realistic Datasets}. 
The three prominent datasets in this space (SARD, HERIDAL, and WiSARD) frequently contain samples that are not realistic to WSAR. 
Targets in these datasets often wear bright, clean clothing, despite missing persons often being covered in mud and dirt when found \cite{koester2008lost}.
Targets are frequently seen holding drone controllers, something that missing persons would likely not possess. 
Targets are also frequently located in open areas or on trails, despite missing persons typically seeking shelter and descending to lower terrain \cite{koester2008lost}.
Finally, objects such as backpacks, clothing, and other evidence of human activity are frequently excluded from datasets, or worse, included intentionally as false positive samples despite being valuable to search efforts.
All of these factors will induce model biases that may result in failures to find missing persons when deployed at scale.

\textbf{Better Models \& Features}. 
The majority of models in the literature are straightforward instantiations of popular CV models that have been trained solely on WSAR imagery.
However, in practice, there are substantial amounts of information that can be used to augment these models.
For example, current models need to manage the scale variance problem described in Section \ref{sec:ops_data} through data augmentation.
But one could imagine learning deformable convolutions \cite{dai2017deformable} conditioned on image resolution, focal length, and flight altitude (similar to the analysis done in \cite{goodrich2008supporting}) to make the scale variances an easier function to learn.
Other models could be trained to leverage details of the missing person, such as age or clothing color. 
Features like these need to be considered when constructing datasets so future models can leverage the data present during real world operations.

\textbf{Alignment on Evaluation Metrics}. 
There are at least five papers that do not report average precision scores for their models.
While, all papers report precision and recall, or statistics that can be used to compute those measures, authors rarely report their model's decision thresholds.
Inconsistent reporting of performance metrics makes it difficult to understand how different approaches compare to one another.
As argued in \cite{pyrro2022air}, this context strongly suggests that the SAR-APD metric is a more appropriate evaluation scheme than VOC2012 \cite{pascal-voc-2012}.
Thus, future papers are encouraged to report Average Precision based on the SAR-APD bounding box evaluation scheme as reported here and in \cite{pyrro2022air}.

\section{Conclusion}
This paper discussed the current state of the art in using sUAS and CV for WSAR; detailed the operational environment, the data collection and the deployment of CV methods on an actual missing persons case; and highlighted three future directions for the research community.


{\small
\bibliographystyle{ieee_fullname}
\bibliography{egpaper_final}
}

\begin{appendices}
\appendix
\section{Details Of Collected Imagery}
\begin{figure}[h!]
\centering
\includegraphics[width=85mm,scale=0.5]{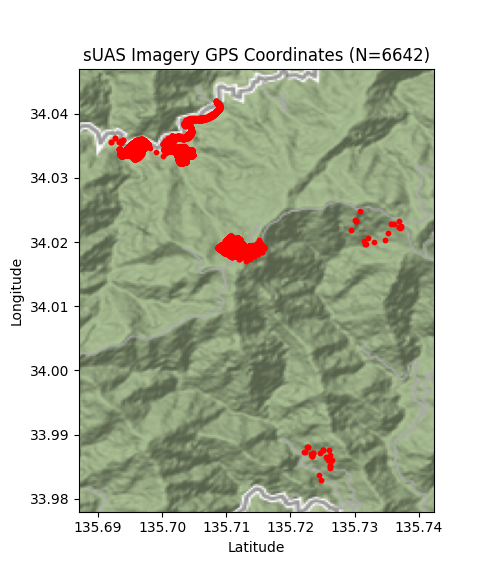}
\caption{The GPS coordinates of imagery collected in the search for Patricia Wu-Murad denoted in red. Imagery that did not have an associated GPS coordinate has been omitted.}
\label{fig:locations}
\end{figure}

\begin{figure}[h!]
\centering
\includegraphics[width=85mm,scale=0.5]{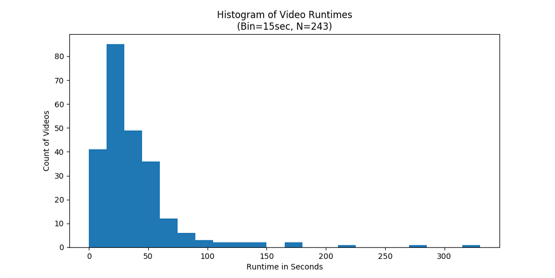}
\caption{Histogram of the runtimes of the 243 collected videos from the search for Patricia Wu-Murad.}
\label{fig:runtimes}
\end{figure}

This section provides supporting details of the sUAS imagery that was collected in support of Patricia Wu-Murad. Figure \ref{fig:locations} contains the geographic information associated with the collected imagery. Figure \ref{fig:runtimes} contains the runtimes of the sUAS videos that were collected during the search.

\section{RX Classifier Hyperparameters}
\label{app:rx}
The following are the hyperparameters used when running the RX spectral classifier model.
\begin{itemize}
    \item \textbf{Image Resize}: All input images are resized to be 1024x1024
    \item \textbf{P-Value}: Pixels are considered anomalous if they have a z-score and associated p-value of $<$ 0.0001
    \item \textbf{DBSCAN eps}: Anomalous pixels were clustered together if their distance was $<$14.4815
    \item \textbf{DBSCAN Min Samples}: Clusters were dropped if they had $<$ 209 pixels. 
    \item \textbf{Cluster Count Limits}: Images that resulted in $<$1 or $>$4 clusters were ignored.    
\end{itemize}

\begin{figure*}[h!]
\centering
\includegraphics[width=\textwidth,scale=0.5]{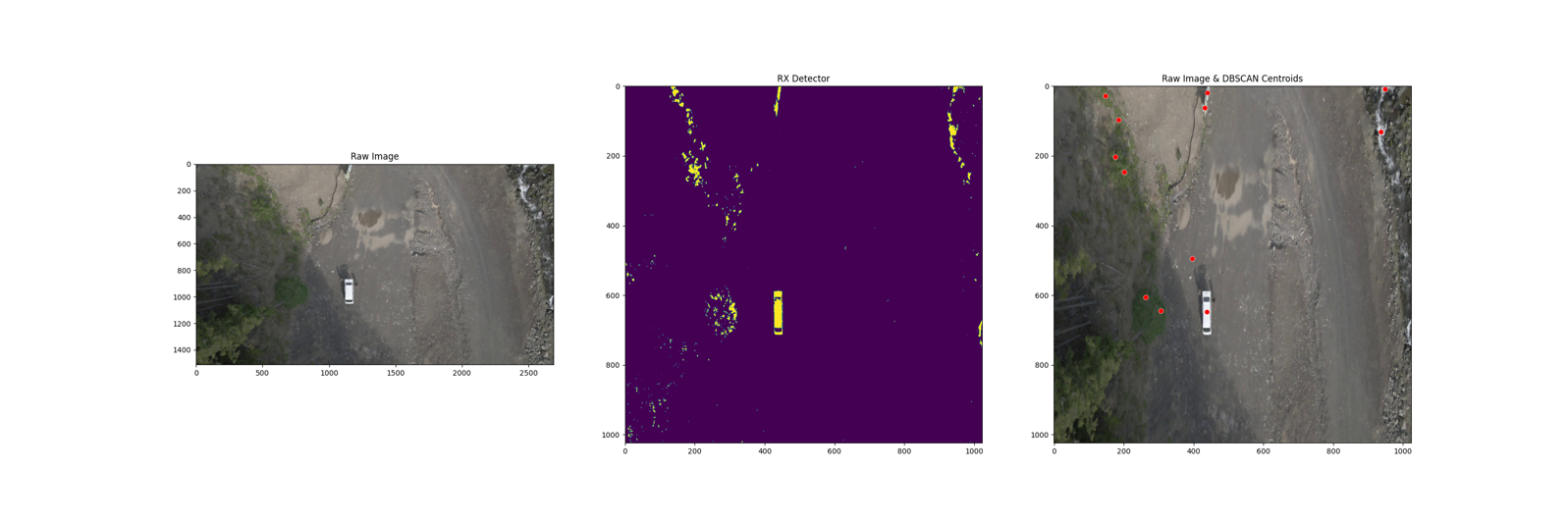}
\includegraphics[width=\textwidth,scale=0.5]{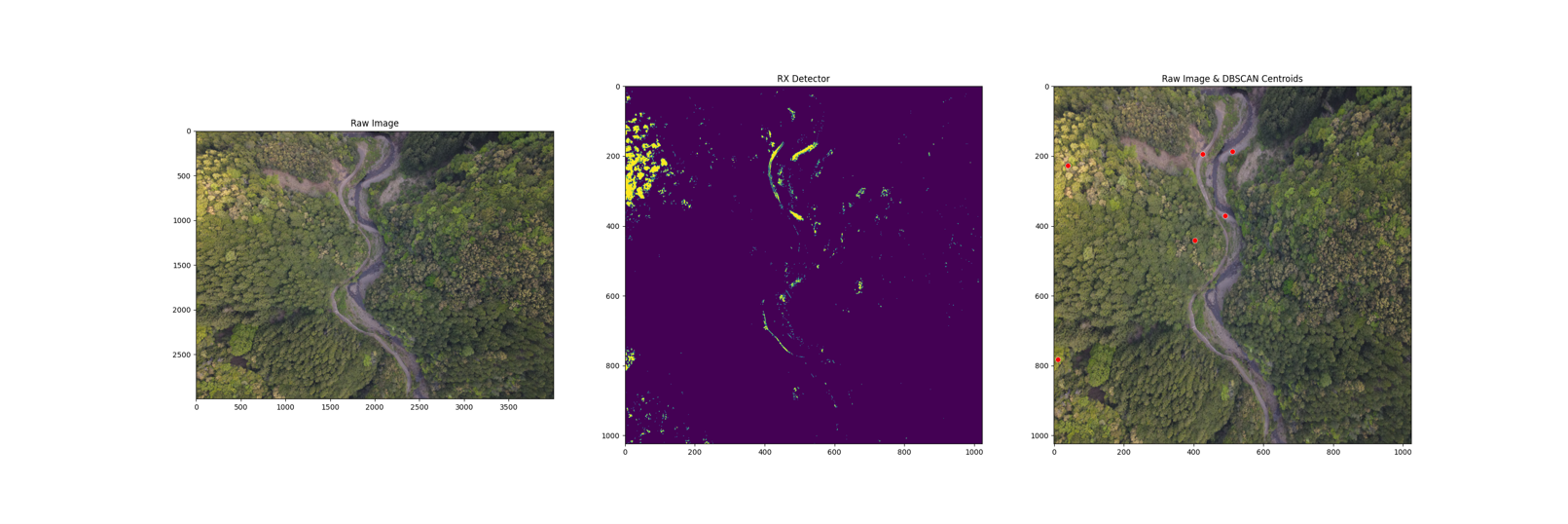}
\caption{Sample inputs and outputs of the Unsupervised RX Spectral classified as applied to imagery collected from the field.}
\label{fig:rx_sample}
\end{figure*}

Sample inputs and outputs, and the intermediate pixel classification step is included in Figure \ref{fig:rx_sample} for awareness.

\section{EfficientDET Training \& Hyperparameters}
\label{app:ed_training}
The input resolution of the EfficientDET model was 512x512 and it was trained on the HERIDAL dataset\cite{maruvsic2018region}.
The tf\_efficientnetv2\_l model \cite{tan2021efficientnetv2} was used as the backbone which was then fine tuned using HERIDAL data.
The HERIDAL has a published train and test set. The train set was randomly split into into 90\% training samples and 10\% validation samples.

Training samples were generated in the following manner. The functionality of the Albumentations library \cite{2018arXiv180906839B} was utilized and extended.
\begin{enumerate}
    \setlength\itemsep{-0.5em}
    \item Select a random image from the HERIDAL training set
    \item Randomly resize the image between 0.7 and 1.1 of the original image size while maintain aspect ratio.
    \item If that image contains no bounding boxes select a random 512x512 crop from the image.
    \item If that image contains at least one bounding box, with equal likelihood
    \setlength\itemindent{15pt} \item[-] Select a random 512x512 crop from that image
    \setlength\itemindent{15pt} \item[-] Select a random bounding box and then select a random 512x512 crop of that image that contains that selected bounding box.
\end{enumerate}

Training samples were augmented in the following manner. The Albumentations library \cite{2018arXiv180906839B} was used to perform these augmentations.
\begin{enumerate}
    \setlength\itemsep{-0.5em}
    \item Randomly flip the image along its x or y axis with 50\% likelihood
    \item Randomly rotate the image at 90 degrees increments between 0-270 around its center
    \item Perform embossing at 25\% likelihood
    \item Synthetically add snow to at 15\% likelihood \cite{AutoMold}
    \item Synthetically add fog to at 15\% likelihood (fog\_coef\_lower=0.01, fog\_coef\_upper=0.4) \cite{AutoMold}
    \item Apply the sepia filter to the image at 10\% likelihood
    \item Add random Gaussian noise to every pixel
    \item Perform one of the following
    \setlength\itemindent{15pt} \item[-] Add random brightness and contrast to the image at 50\% likelihood
    \setlength\itemindent{15pt} \item[-] Jitter the color of the image at 50\% likelihood
    
\end{enumerate}

Finally, the color values of the image were normalized according to the following color channel mean values [R: 0.485, G: 0.456, B: 0.406] and standard deviation values [R: 0.229, G: 0.224, B: 0.225].

The default set of hyperparameter set forth in the EffDet python library \cite{effdet} were used. The relevant hyperparameters are detailed below.
\begin{itemize}
    \setlength\itemsep{-0.5em}
    \item \textbf{Learning Rate}: 0.0002
    \item \textbf{Optimizer}: Adam
    \item \textbf{backbone\_drop\_path\_rate}: 0.2
	\item \textbf{backbone\_indices}: None
	\item \textbf{min\_level}: 3
	\item \textbf{max\_level}: 7
	\item \textbf{num\_levels}: 5
	\item \textbf{num\_scales}: 3
	\item \textbf{anchor\_scale}: 4.0
	\item \textbf{pad\_type}: same
	\item \textbf{act\_type}: swish
	\item \textbf{norm\_layer}: None
	\item \textbf{norm\_eps}: 0.001
	\item \textbf{norm\_momentum}: 0.01
	\item \textbf{box\_class\_repeats}: 3
	\item \textbf{fpn\_cell\_repeats}: 3
	\item \textbf{fpn\_channels}: 88
	\item \textbf{separable\_conv}: True
	\item \textbf{apply\_resample\_bn}: True
	\item \textbf{conv\_bn\_relu\_pattern}: False
	\item \textbf{downsample\_type}: max
	\item \textbf{upsample\_type}: nearest
	\item \textbf{redundant\_bias}: True
	\item \textbf{head\_bn\_level\_first}: False
	\item \textbf{head\_act\_type}: None
	\item \textbf{fpn\_name}: None
	\item \textbf{fpn\_config}: None
	\item \textbf{fpn\_drop\_path\_rate}: 0.0
	\item \textbf{alpha}: 0.25
	\item \textbf{gamma}: 1.5
	\item \textbf{label\_smoothing}: 0.0
	\item \textbf{legacy\_focal}: False
	\item \textbf{jit\_loss}: False
	\item \textbf{delta}: 0.1
	\item \textbf{box\_loss\_weight}: 50.0
	\item \textbf{soft\_nms}: False
	\item \textbf{max\_detection\_points}: 5000
	\item \textbf{max\_det\_per\_image}: 100
\end{itemize}

The bounding boxes that are generated by the model are fused together using the Weighted Box Fusion technique described in \cite{solovyev2021weighted}. 
When performing tiled inference on the HERIDAL dataset, or on data collected operationally, any bounding boxes that are overlapping are merged together by generating a larger bounding box that contains exactly the two overlapping bounding boxes.

The \textit{Tiled\_EffecientDET$_{84}$} and \textit{Tiled\_EffecientDET$_{174}$} models were trained for 84 and 174 epochs respectively, where one epoch was defined as having generated one sample from each image in the training set. The training process for \textit{Tiled\_EffecientDET$_{174}$} took approximately 16 hours and training was conducted on a Nvidia 1080TI. 

Inference on the 18759 images generated in the search for Patricia Wu-Murad took approximately 4.74 hours and on average 0.91 images per sec were processed. 

\section{Four Candidate Regions of Interest}
\label{app:cand}
\begin{figure*}[h!]
\centering
\includegraphics[width=\textwidth,scale=0.5]{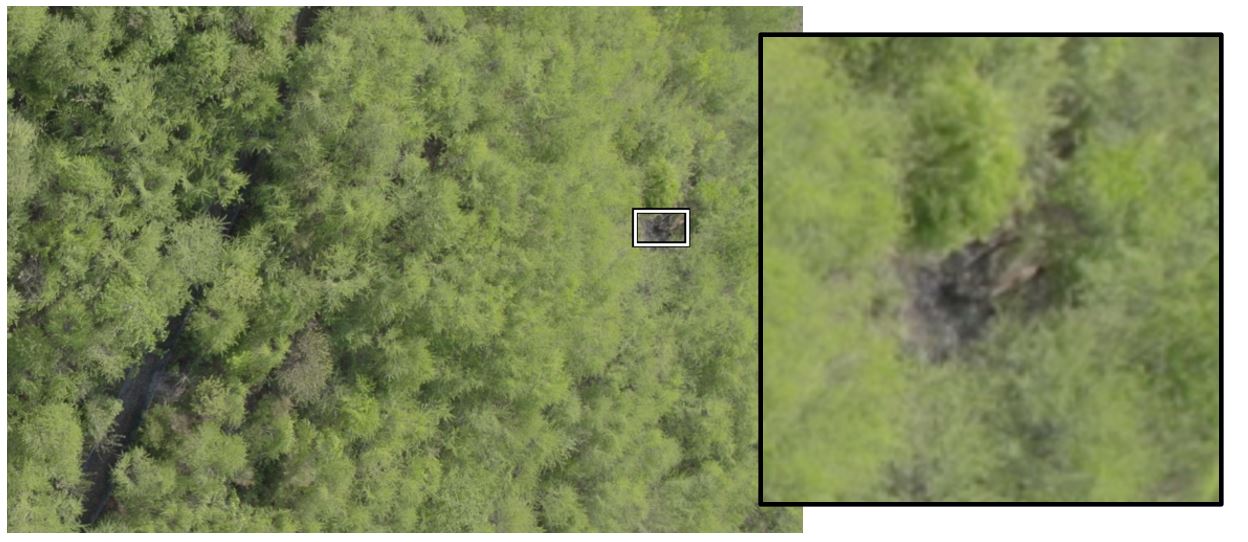}
\caption{Candidate Image \#1 that was handed off to the ground search teams.}
\label{fig:cand_1}
\end{figure*}

\begin{figure*}[h!]
\centering
\includegraphics[width=\textwidth,scale=0.5]{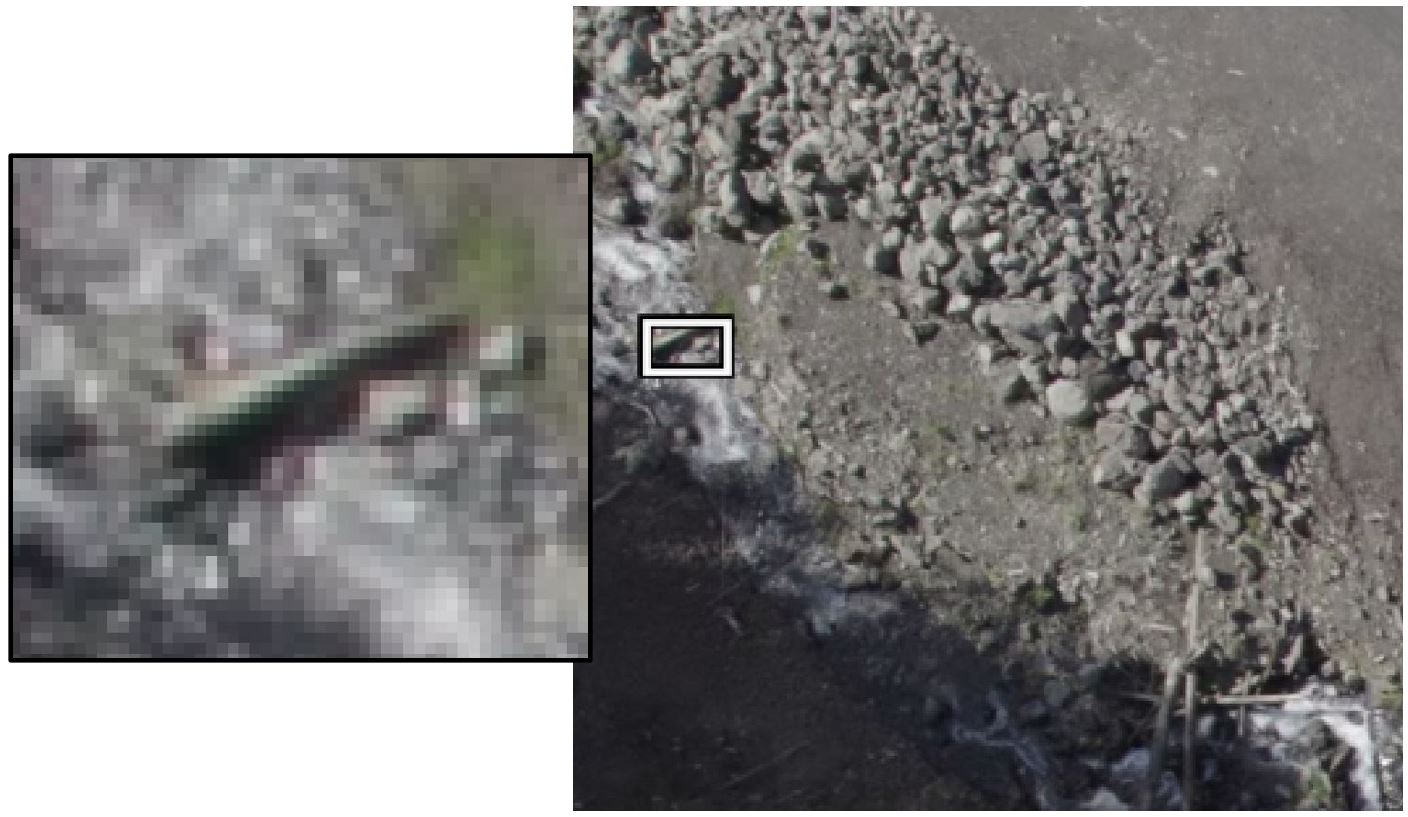}
\caption{Candidate Image \#2 that was handed off to the ground search teams.}
\label{fig:cand_2}
\end{figure*}

\begin{figure*}[h!]
\centering
\includegraphics[width=\textwidth,scale=0.5]{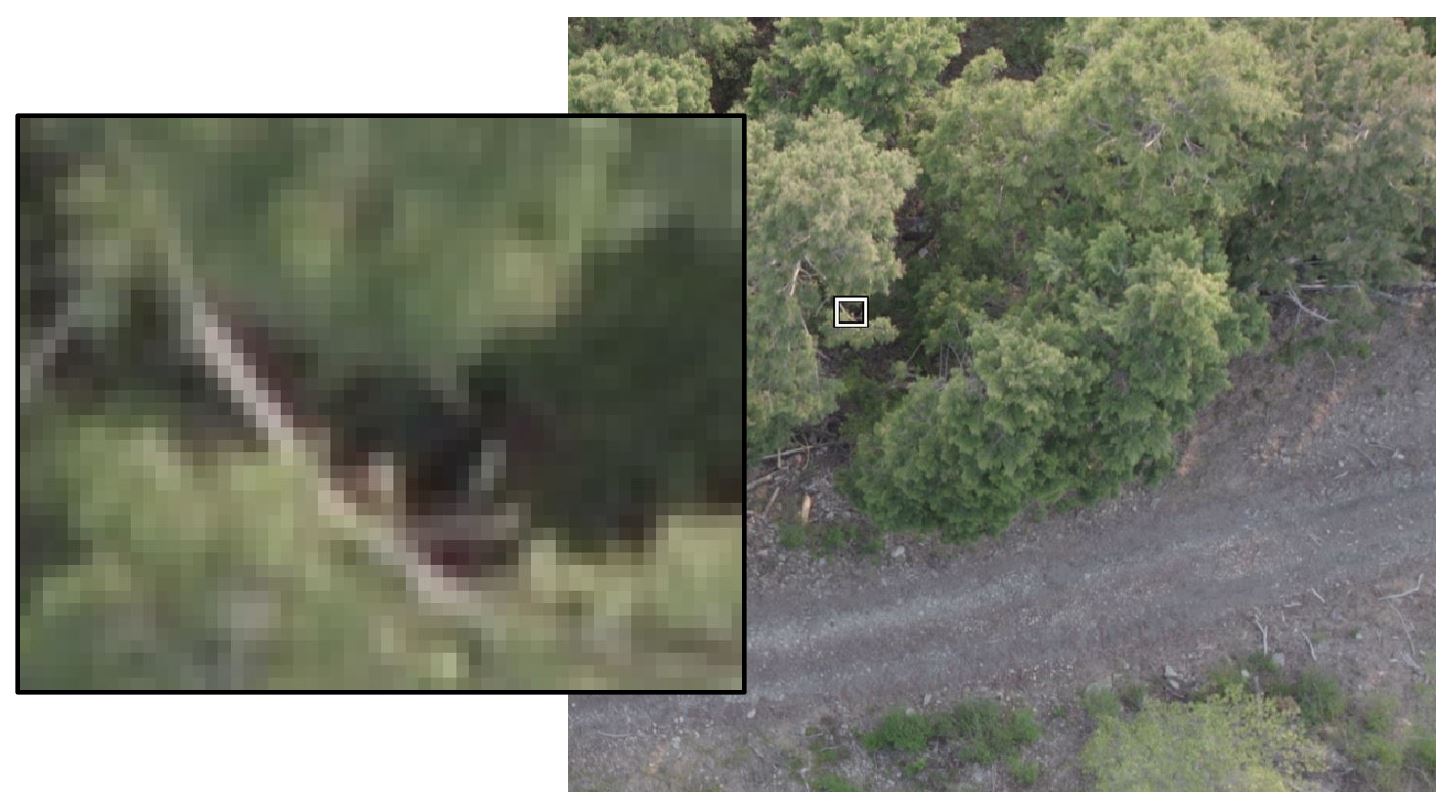}
\caption{Candidate Image \#3 that was handed off to the ground search teams.}
\label{fig:cand_3}
\end{figure*}

\begin{figure*}[h!]
\centering
\includegraphics[width=\textwidth,scale=0.5]{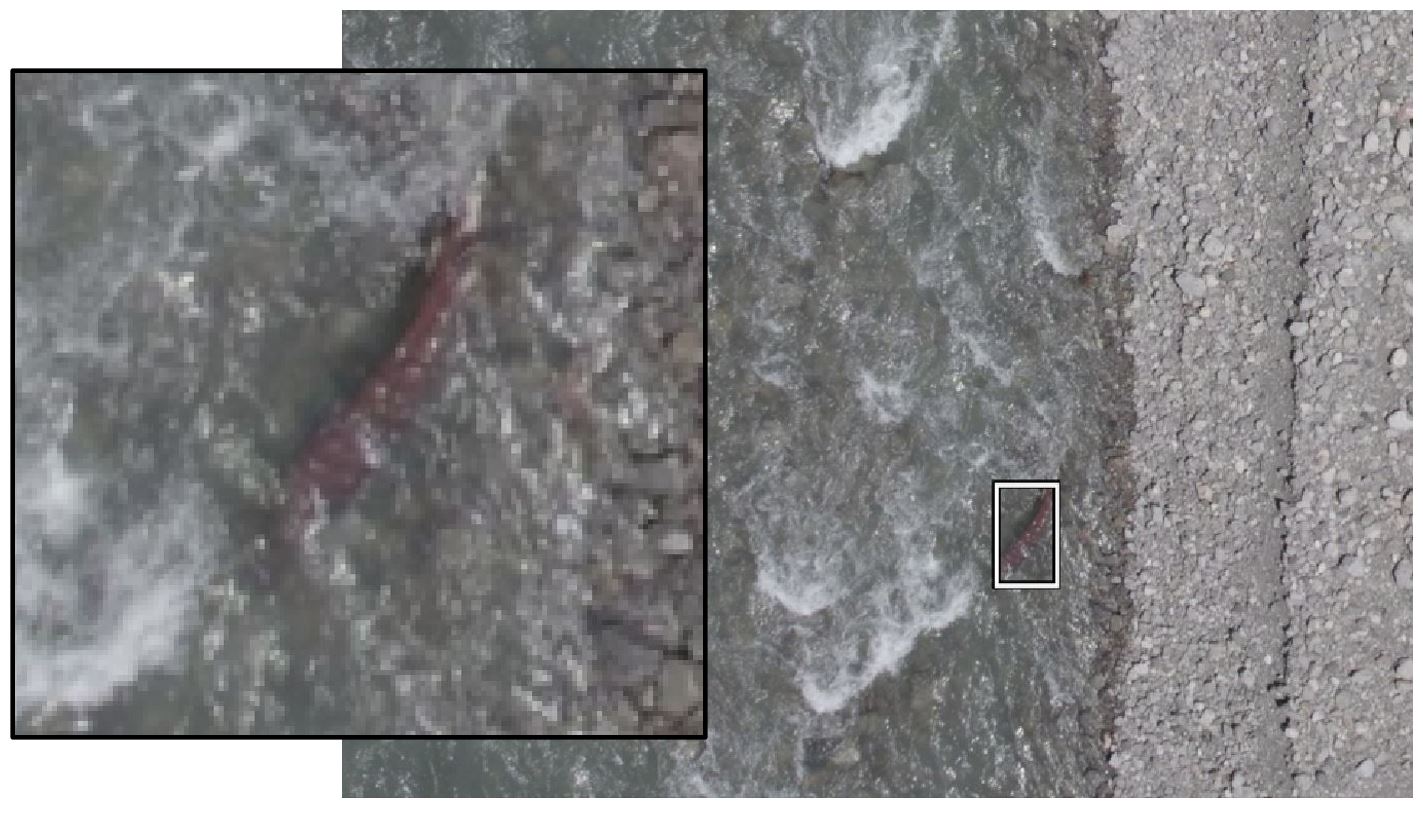}
\caption{Candidate Image \#4 that was handed off to the ground search teams.}
\label{fig:cand_4}
\end{figure*}

The four candidate regions of interest that were handed off to the ground search teams are contained in Figures \ref{fig:cand_1}, \ref{fig:cand_2}, \ref{fig:cand_3}, and \ref{fig:cand_4}. None of these regions of interest appear to show a person, instead they show an abnormality that cannot be immediately explained.

\end{appendices}

\end{document}